%% file: revised_speech.tex
\documentclass[11pt]{article}

\usepackage{acl}
\usepackage{times}
\usepackage{latexsym}
\usepackage[T1]{fontenc}
\usepackage[utf8]{inputenc}
\usepackage{microtype}
\usepackage{inconsolata}
\usepackage{graphicx}
\usepackage{amsmath,amssymb}
\usepackage{booktabs,array}
\usepackage{placeins}
\newcolumntype{L}[1]{>{\raggedright\arraybackslash}p{#1}}
\newcommand{\pmv}[2]{#1\,$\pm$\,#2}
\title{MedAstra: ASR-less Speech Retrieval with Diagnostic Grounding for Medical QA}
\author{Elahe Khatibi, Ziyu Wang, Amir M. Rahmani}

\begin{document}
\maketitle
\begin{abstract}
Spoken medical question answering can fail when automatic speech recognition (ASR) corrupts a clinically important term before retrieval. We present \textsc{MedAstra}, a framework that maps frozen WavLM speech representations through a lightweight adapter into a frozen E5 text embedding space. This enables a dense retrieval path that accepts a waveform without a query transcript. Retrieved passages and speech-linked textual metadata support answer generation with entity alternatives, diagnostic contrasts, relation heuristics, and bounded retrieval refinement. ASR-less refers to the initial speech retrieval interface; lexical retrieval, string-based entity correction, and the text generator require textual inputs. The deterministic speech embedding does not constitute a calibrated model of acoustic uncertainty. Reported retrieval results reach 87.6 Recall@10 for the speech-only retriever and 90.4 for the hybrid configuration, compared with 81.0 for transcript-first dense retrieval. Medical QA and ablations provide descriptive system comparisons, with input-provenance and evaluation limitations stated explicitly. We distinguish controlled synthetic acoustic tests from real clinical speech, qualify component attribution, and specify the remaining comparisons needed to assess speech-only QA and clinical generalization.
\end{abstract}

\section{Introduction}
\label{sec:intro}
\begin{figure*}[t]
\centering
\includegraphics[width=0.95\textwidth]{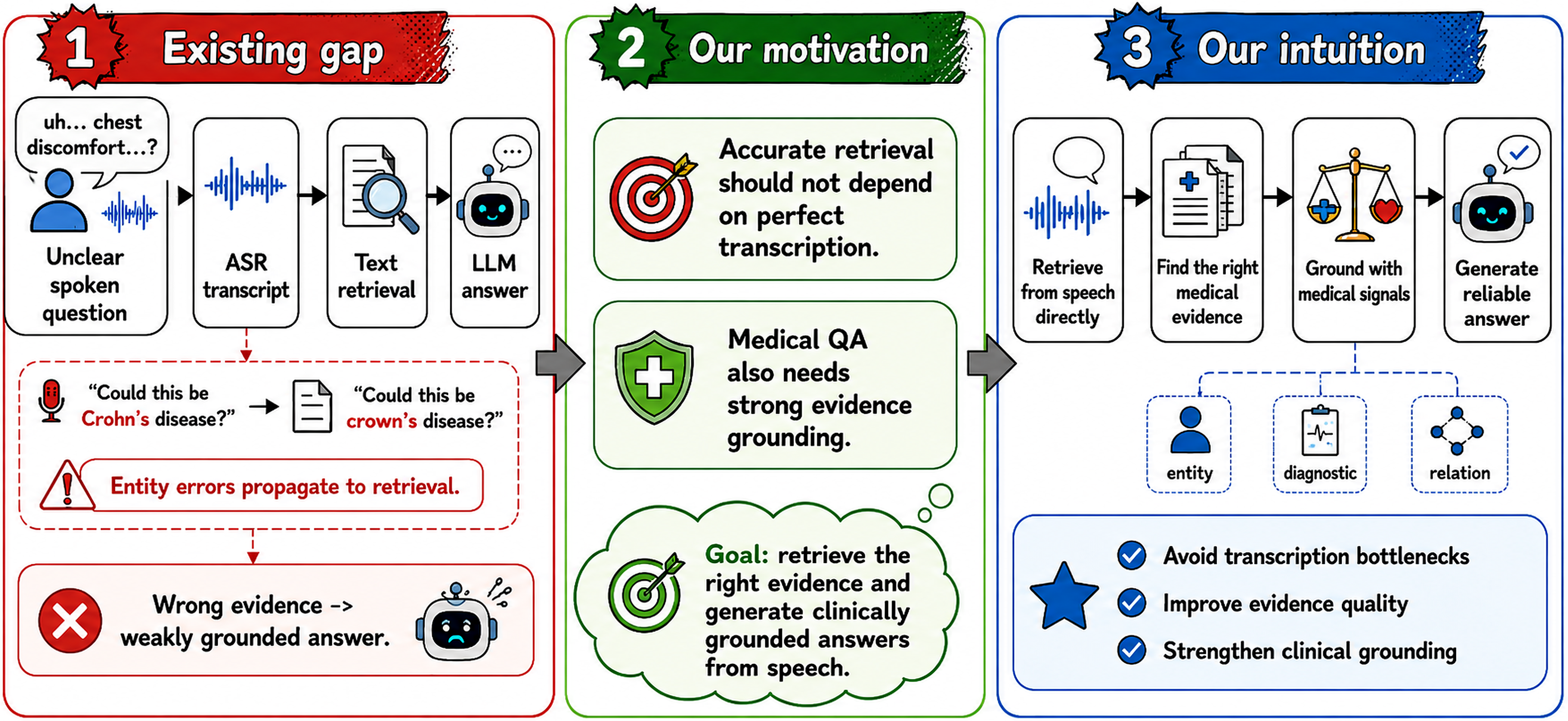}
\caption{Motivation for waveform-based medical evidence retrieval. Crohn's/crown's illustrates a possible ASR substitution, not a recorded benchmark case. The right panel depicts design goals, not established clinical reliability.}
\label{fig:motivation}
\end{figure*}

Speech offers an accessible interface to medical information, but transcript-first retrieval introduces a consequential decision before evidence selection. An ASR system may replace a rare entity, abbreviation, or near-homophone with a familiar term, after which a text retriever searches for the wrong concept. Clinical transcription errors and their effects on retrieval motivate examining this interface separately from final-answer accuracy \citep{zhou2018analysis,sidiropoulos2022impact,wang2020asr}. A fluent answer cannot recover information that is absent from its evidence. Figure~\ref{fig:motivation} illustrates this motivation.

\textsc{MedAstra} connects speech-to-evidence alignment with medical evidence grounding. Frozen WavLM features \citep{chen-etal-2022-wavlm} are projected into a frozen E5 space \citep{wang-etal-2022-e5}; the speech embedding can search both aligned speech items and text passages. A text-based generator consumes readable evidence and optional entity, diagnostic, and relation context. The modular design separates the acoustic retrieval interface from downstream generation.

Two boundaries are essential. First, bypassing a query transcript is not equivalent to eliminating text throughout the system. Speech-linked evidence requires associated text for the current generator, while BM25, text reranking, and edit distance require a text query. Second, a single embedding and a ranked evidence set do not model or calibrate acoustic uncertainty. We therefore describe the system as retaining an acoustic retrieval path and exposing alternative evidence, without claiming probabilistic uncertainty preservation.

Our contributions are (i) a lightweight speech-to-evidence interface between frozen encoders; (ii) an explicit contract between dense speech retrieval and text-dependent grounding; and (iii) an analysis of reported retrieval, acoustic robustness, medical QA, and ablation results with their evidential limits. The study does not establish a fully speech-only medical QA pipeline, clinical safety, or generalization to patient populations. Its clearest empirical question is whether the speech retrieval path can recover evidence that a single ASR transcript misses.

Medical QA also requires evidence that preserves the queried entity and distinguishes plausible answers. A passage about a similar disease may be topically relevant yet lack the decisive contrast. We therefore examine three decisions: which evidence the waveform retrieves, which entity and diagnostic alternatives reach the prompt, and whether the answer is supported by that visible context. Acoustic retrieval misses, inappropriate term replacements, and unsupported generated claims require different remedies. The targeted entity and diagnosis suites complement broad QA benchmarks by examining these dependencies.

\section{Related Work}
\label{sec:related}
\paragraph{Speech retrieval.}
SpeechRAG retrieves speech passages from text queries and uses a speech language model for generation \citep{min-etal-2025-speechrag}. Our SpeechRAG-style comparison is an adapted alignment baseline, not the original system or its published scores. SEAL aligns speech and text representations for retrieval \citep{sun-etal-2025-seal}. WavLM, HuBERT, and wav2vec~2.0 provide alternative speech encoders \citep{chen-etal-2022-wavlm,hsu-etal-2021-hubert,baevski-etal-2020-wav2vec2}. SONAR supplies pretrained speech and text encoders in a shared sentence embedding space \citep{duquenne-etal-2023-sonar}; it is a relevant alternative to training a WavLM--E5 adapter. An encoder swap would also require rebuilding the index in the replacement text space. We evaluate one encoder pair and do not infer superiority over joint encoders.

\paragraph{Medical retrieval and grounding.}
Dense retrieval and RAG condition generation on external evidence \citep{karpukhin-etal-2020-dense,lewis-etal-2020-rag}. MedRAG adds graph-elicited diagnostic reasoning \citep{zhao-etal-2025-medrag}. Medical spoken entity recognition and ASR correction address term preservation \citep{le-duc-etal-2025-medical,luo-etal-2025-generative}, while contextual ASR uses retrieved context during transcription \citep{xiao-etal-2025-contextual-asr}. These approaches complement speech retrieval; they are not ruled out by a comparison with a single-best-transcript baseline. ASR n-best fusion, confidence-aware retrieval, and entity-corrected transcripts remain necessary stronger controls.

\paragraph{Retrieval refinement.}
Active retrieval and Self-RAG motivate revisiting evidence after generation \citep{jiang-etal-2023-active,asai-etal-2024-selfrag}. Our bounded loop uses simple generated-term feedback. The diagnostic graph and relation module provide heuristic context, rather than causal identification or a validated clinical reasoner. In particular, test--diagnosis links are evidential associations and should not be read as causal effects.

\section{Method}
\label{sec:method}
\subsection{Problem Setup and Module Interfaces}
\label{sec:overview}
For a spoken question $x$, the retrieval stage selects a compact set $E$ from speech-linked items $A$ and text passages $D$. Each item has an identifier, a retrieval vector, and readable metadata. The answering stage receives that metadata, a textual question when available, and the task's answer choices. Its output $y$ can be an option label or an open answer, depending on the evaluation contract.

\begin{figure*}[t]
\centering
\includegraphics[width=0.95\textwidth]{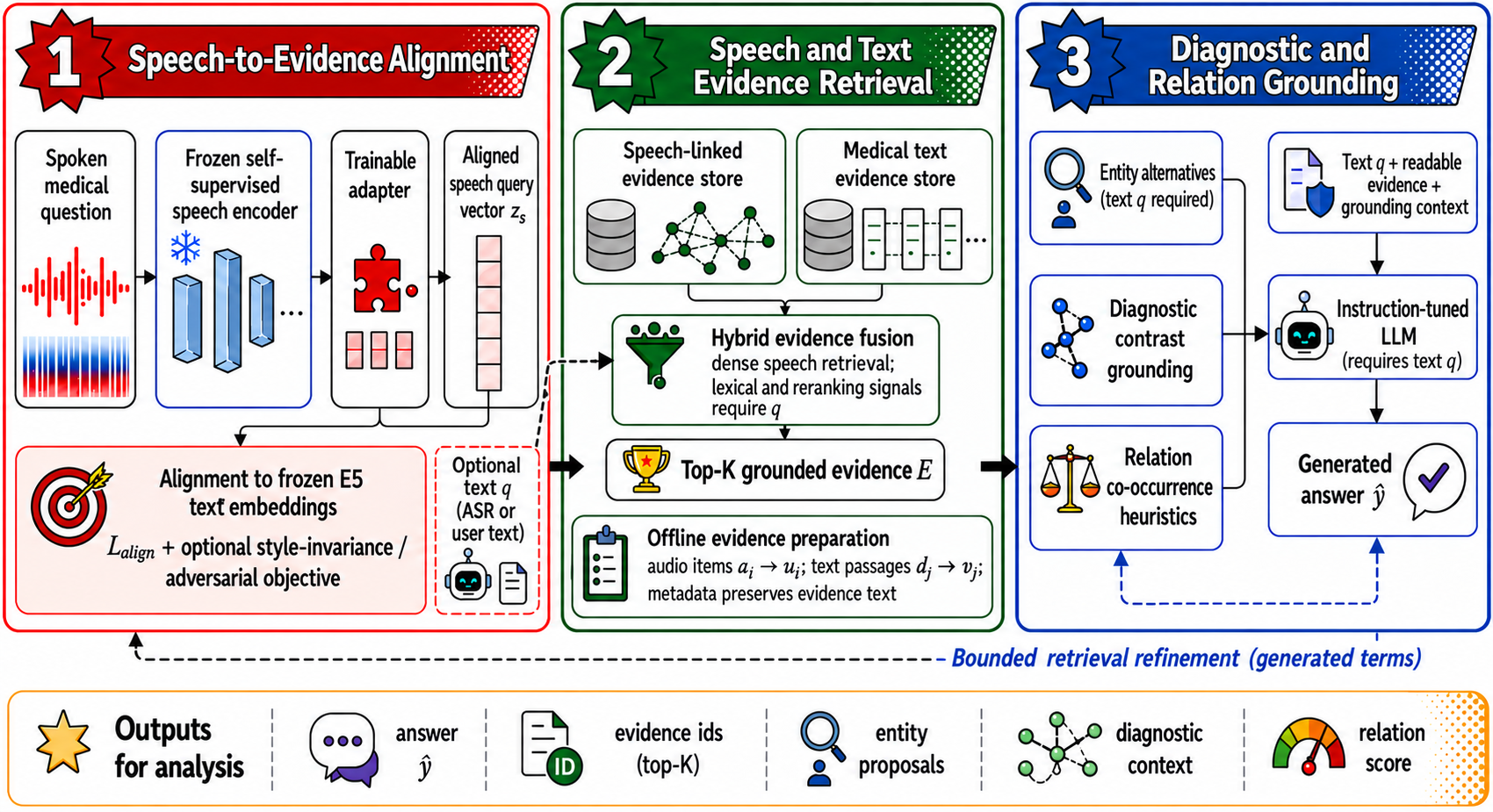}
\caption{\textsc{MedAstra}'s three-stage design. The waveform drives dense retrieval; lexical signals, reranking, string correction, and generation require text. Entity alternatives, diagnostic contrasts, and relation heuristics supply context without guaranteeing answer support.}
\label{fig:method}
\end{figure*}

Alignment, evidence selection, and answer synthesis have distinct inputs: paired speech/reference text for alignment training, a query waveform for dense retrieval, and text for string-based grounding and current generation. These interfaces allow acoustic retrieval to be assessed independently of the answering text channel (Figure~\ref{fig:method}).

\subsection{Inputs and Speech-to-Evidence Alignment}
\label{sec:alignment}
Let $x$ be a waveform, $q$ an optional text query, $A$ a collection of speech items with readable metadata, and $D$ a text passage collection. A reference transcription $t$ is used for paired training, not required by the dense speech query at inference. We pool frozen WavLM states, project them through an adapter $g_\theta$, and normalize:
\begin{equation}
 \begin{aligned}
 z_s&=\mathrm{norm}(g_\theta(\mathrm{Pool}(f_s(x)))),\\
 z_t&=\mathrm{norm}(f_t(t)).
 \end{aligned}
\end{equation}
Both encoders remain frozen. The adapter consists of two 768-dimensional linear layers, ReLU, dropout, and output layer normalization. Alignment minimizes $\mathcal L_{\mathrm{align}}=1-z_s^\top z_t$. Normalized vectors are searched using inner product in FAISS \citep{johnson-etal-2021-faiss}.

Waveforms are converted to mono at 16 kHz and truncated or zero-padded to an 8-second window. The wrapper uses mean pooling, with masked pooling only when the available mask matches the encoder's frame sequence. E5 uses query and passage prefixes for the respective retrieval roles. Freezing the encoders fixes the destination space while training the adapter; it simplifies replacing the speech interface, but does not remove index, pooling, or prompt differences between systems. Exact preprocessing and model revisions remain part of each experiment's configuration.

An optional augmented view $\tilde{x}_i$ supplies an invariance objective:
\begin{equation}
\mathcal L_{\mathrm{inv}}=-\log\frac{\exp(z_i^\top\tilde z_i/\tau)}{\sum_j\exp(z_i^\top\tilde z_j/\tau)}.
\end{equation}
The implementation averages this contrastive term in both view directions. A gradient-reversal head is intended to suppress augmentation information \citep{oord-etal-2018-cpc,ganin-etal-2016-domain-adversarial}; its current feature/label wiring requires verification (Appendix~\ref{app:settings}). The combined loss is
\begin{equation}
\mathcal L=\lambda_a\mathcal L_{\mathrm{align}}+
\lambda_i\mathcal L_{\mathrm{inv}}+
\lambda_v\mathcal L_{\mathrm{adv}}.
\end{equation}
Code defaults are $\tau=0.07$, $(\lambda_a,\lambda_i,\lambda_v)=(1,0.2,0.1)$, augmentation probability 0.5, and reversal coefficient 1.0. These are recoverable implementation settings; historical run configurations still require verification (Appendix~\ref{app:settings}). Neither objective yields a posterior over spoken terms.

\subsection{Dense Retrieval and Text-Dependent Fusion}
\label{sec:fusion}
Speech items are indexed as $u_i=\mathrm{norm}(g_\theta(\mathrm{Pool}(f_s(a_i))))$ and text passages as $v_j=\mathrm{norm}(f_t(d_j))$. With no text input, $z_s$ can query both dense stores. A speech hit is passed to the generator through its stored reference text or description; the generator does not interpret raw retrieved waveforms. This metadata requirement differs from a query-time ASR requirement.

When $q$ is available, BM25 \citep{robertson-zaragoza-2009-bm25} adds lexical candidates and a text cross-encoder can rerank. Their query source must be recorded as user text, ASR output, reference text, or generated feedback. Reference text is an oracle condition. It must not silently enter a speech-only result. Without $q$, first-round BM25, query-string entity correction, and text-pair reranking are disabled. These modules cannot consume $z_s$ as if it were text.

Reciprocal-rank fusion combines available rankings \citep{cormack-etal-2009-rrf}:
\begin{equation}
 S(d)=\sum_{\ell\in\mathcal R}\frac{1}{60+\mathrm{rank}_{\ell}(d)}.
\end{equation}
The intended fusion sorts candidates by $S(d)$ before optional reranking and retains source IDs. The current medical runner computes RRF scores but does not use them to order its cross-encoder candidate list; it reranks a dense/lexical union. We distinguish this implementation from the intended rank-fusion contract instead of treating them as experimentally equivalent.

The evidence union preserves document identifiers so repeated hits can be merged. Candidate scores from different stores are not calibrated probabilities; RRF combines their ranks rather than directly comparing score magnitudes. Text-dependent modules can improve exact-term matching when a textual channel exists, but adding them changes the condition being tested. Appendix Table~\ref{tab:input-contract} makes this distinction explicit for paired-text benchmarks and for waveform-only retrieval.

\subsection{Entity, Diagnostic, and Relation Context}
\label{sec:grounding}
The entity corrector ranks a textual token $e$ against a medical vocabulary. The phonetic feature $s_p$ is Soundex equality, the edit feature $s_e=1/(1+\mathrm{Lev}(e,c))$, and $s_m$ is a semantic-neighbor feature:
\begin{equation}
 \begin{split}
 s(e,c)={}&0.55s_p(e,c)+0.35s_e(e,c)\\
 &+0.10s_m(e,c).
 \end{split}
\end{equation}
The replacement threshold is strictly greater than 0.75; up to three proposals are returned for retrieval-query correction. Scores are heuristic similarities, not probabilities. Without semantic neighbors, a different term has score at most $0.55+0.35/2=0.725$, so the default corrector proposes alternatives but cannot replace it at that threshold. Consequently, entity correction improvements cannot be attributed to this implementation without identifying the actual vocabulary, semantic context, and run configuration.

If no query string exists, phonetic and edit comparisons against the unknown spoken token are undefined. Terms from retrieved metadata can be shown as evidence-derived alternatives, but these do not constitute recognition of the original spoken entity. Post-retrieval proposals cannot repair a first-stage retrieval miss unless later retrieval actually consumes them.

For diagnostic contrast, let $O$ be the extracted manifestation set and $S_d$ the graph's symptom set for disease $d$. The available graph ranks candidates by Jaccard overlap:
\begin{equation}
 J(d,O)=\frac{|O\cap S_d|}{\max(1,|O\cup S_d|)}.
\end{equation}
For the top two candidates, it supplies shared features $S_{d_1}\cap S_{d_2}$ and distinguishing sets $S_{d_1}\setminus S_{d_2}$ and $S_{d_2}\setminus S_{d_1}$. Each list is capped at 20 entries. Missing graph features can become follow-up prompts; the QA runner includes at most three for the first candidate. These are suggested questions, not observed patient findings. Its token extractor is a lexical heuristic, so a matching token does not establish a symptom's presence or interpret negation.

The relation module forms candidate disease--symptom, treatment--outcome, and test--diagnosis edges from keyword co-occurrence and fixed type priors. Co-occurrence does not establish entailment. The available checker rewards endpoint mentions and penalizes a literal reversal pattern:
\begin{equation}
 \begin{split}
 s_{\mathrm{rel}}(y,G)={}&0.2\sum_{(u,v,c)\in G}c\,\mathbb I[u,v\in y]\\
 &-0.5\,\mathbb I[\mathrm{pattern}(y)].
 \end{split}
\end{equation}
Its pattern matches ``symptom ... causes ... disease.'' This raw heuristic is neither a normalized percentage nor a general detector of relation direction. We do not interpret it as causal consistency, causal discovery, or clinical safety validation.

\subsection{Generation and Bounded Refinement}
\label{sec:iteration}
The generator receives a textual question, choices when applicable, readable evidence, and available grounding context. The available QA runner initializes from \texttt{question\_text}; a complete question representation derived solely from the waveform is not implemented there. Thus ASR-less initial retrieval and text-dependent answer synthesis must be assessed separately.

At most two rounds are used by default. The implemented text refinement is
\begin{equation}
 q^{(r+1)}=\mathrm{Correct}(q^{(0)}\oplus\mathrm{Terms}(y^{(r)})),
\end{equation}
where $\mathrm{Terms}$ extracts up to eight alphabetic tokens of length at least four and $\mathrm{Correct}$ is enabled only in the entity-correction condition. The runner corrects the query before initial retrieval and again after feedback. Post-answer proposal objects are stored in the trace; they are not explicitly appended to the next query. The answer schema \texttt{Correct Option: A} contributes no clinical entity, and formatting words are not meaningful refinement cues. The existing no-iteration row tests the full system at one round, not the independent effect of entity proposals; the missing entity-on/off comparison at one round remains necessary.

Stopping uses a best retrieved score improvement below 0.01 after the second retrieval and generation. Because the query changes between rounds and stopping follows generation, this heuristic is not evidence of monotonic relevance improvement and need not avoid the second generation cost.

\paragraph{Prompt construction and audit.}
The compact prompt presents the question and choices before the evidence blocks. The local builder takes up to four text snippets and two speech-linked metadata snippets, each capped at 200 characters, within a 1,024-token input limit. The QA-v2 runner currently populates only text hits and inserts diagnostic contrasts and up to eight relation edges into its question block. Its default output budget is 24 tokens with greedy decoding. Thus a final retrieval budget of eight does not mean that all eight items are visible to the generator. Evidence-support evaluation must inspect the exact assembled prompt after truncation. Appendix~\ref{app:prompts} gives distinct answer contracts for exam QA, PubMedQA, and open answers.

\section{Experimental Scope and Protocol}
\label{sec:experiments}
\paragraph{Tasks.}
LibriSpeech supplies paired speech/text alignment data \citep{panayotov-etal-2015-librispeech}. The supplied manuscript reports speech retrieval and additive-noise experiments on SpokenSQuAD \citep{lee-etal-2018-spoken-squad}. Its public release contains Google-TTS passages and textual questions, not recorded spoken medical queries. The waveform-query adaptation and relevance mapping used for the reported retrieval tests require separate provenance. The label ``VoxPopuli-Med'' denotes an author-derived subset, not a public clinical benchmark. VoxPopuli itself contains parliamentary speech \citep{wang-etal-2021-voxpopuli}; the retained subset IDs and medical selection rule are not available. We therefore make no clinical-accent or speaker-group result claim from it.

Medical QA uses PubMedQA, MedQA-USMLE, and MedMCQA \citep{jin-etal-2019-pubmedqa,jin-etal-2021-medqa,pmlr-v174-pal22a}. The stated construction synthesizes question audio while retaining textual choices and labels. The QA labels remain those of the source text benchmarks. MedSpeak-Confusion and Diagnostic-DiffQA are author-derived stress tests, not population samples; their construction and split manifests require completion (Appendix~\ref{app:data}). Additive noise at 20, 10, and 5 dB tests a controlled perturbation, not accents, spontaneous disfluencies, pronunciation variation, or clinical interaction.

\paragraph{Baseline roles and inference settings.}
The retrieval comparisons cover lexical transcript retrieval, Whisper plus E5 dense retrieval, adapted SpeechRAG/SEAL alignment, and MedAstra speech and hybrid configurations. QA comparisons include an LLM-only control, text RAG, contextual-ASR retrieval, and MedRAG-style graph context. These baselines probe different pipeline choices; their names do not establish equality of downstream modules. The available Whisper baseline defaults to large-v3 with beam size five, which is not a stored n-best retrieval condition.

Recoverable MedAstra settings use 50 dense and 50 lexical candidates, eight final hits, BM25 upstream defaults $k_1=1.5$ and $b=0.75$, RRF constant 60, and the MiniLM-L6-v2 reranker. The development adapter configuration uses AdamW, learning rate $2\times10^{-4}$, batch eight, weight decay 0.01, and clipping 1.0. Appendix~\ref{app:settings} distinguishes these defaults from missing historical run settings, including checkpoint selection, synthesis voices, model revisions, and compute budget.

\paragraph{Input and result provenance.}
This revision retains reported retrieval and answer-accuracy values; it adds no new benchmark runs. The supplied original describes three runs, whereas the intermediate revision specifies five without changed numerical cells. The actual seeds and run counts are not recoverable from the available artifacts. We retain the supplied means and standard deviations as descriptive reported statistics, without significance claims. The medical QA runner consumes reference question text, and the Whisper baseline can use that text instead of its transcript when provided. The query-source provenance of the reported tables is therefore unresolved. QA scores must not be read as verified speech-only performance or a transcript-versus-speech causal comparison.

\paragraph{Comparison scope.}
Table~\ref{tab:qa} compares complete configurations: fusion, reranking, correction, graph context, and refinement can differ. A fixed generator alone does not isolate the query interface. Table~\ref{tab:ablation} is the reported component-removal study; its controls and input provenance require confirmation, and its largest drop is an association within that configuration. Decoder swaps test absolute performance within \textsc{MedAstra}; without repeating baselines for each decoder, they do not show that method rankings transfer across decoder families. Gold-text RAG is an input oracle, not a theoretical upper bound on differently augmented systems.

\paragraph{Metrics.}
Retrieval uses target membership in the top $k$ ranks. Accuracy requires a valid task label; for PubMedQA these are \texttt{yes/no/maybe}, while exam QA uses its actual option set. Missing or malformed answers count as incorrect. PubMedQA must not be parsed by an A--D-only rule. We withhold the previous numerical faithfulness and relation-consistency columns because the evidence judge, adjudication records, and percentage conversion are not documented. Answer correctness and evidence support are different metrics: a correctness judge that receives a gold answer but no retrieved evidence cannot assess faithfulness. Appendix~\ref{app:evaluation} specifies a prospective support-evaluation protocol, not a retroactive explanation of the withheld values.

\section{Reported Results and Interpretation}
\label{sec:results}
\subsection{Retrieval and Acoustic Perturbations}
\label{sec:retrieval-results}
Table~\ref{tab:retrieval} reports 87.6 Recall@10 for the speech-only \textsc{MedAstra} row, 6.6 points above Whisper plus dense retrieval, and 90.4 for the hybrid row. The speech-only row also exceeds SpeechRAG-style and SEAL-style rows by 3.3 and 2.6 points. These are adapted method comparisons with the provenance limits in Section~\ref{sec:experiments}; they do not establish that eliminating ASR alone causes the difference.

\begin{table}[t]
\centering\footnotesize\setlength{\tabcolsep}{5pt}
\begin{tabular}{@{}lrrr@{}}
\toprule
Method & R@1 & R@5 & R@10\\
\midrule
BM25 transcript & \pmv{41.8}{0.6} & \pmv{63.4}{0.7} & \pmv{72.1}{0.8}\\
Whisper + dense & \pmv{52.3}{0.5} & \pmv{73.5}{0.6} & \pmv{81.0}{0.7}\\
SpeechRAG-style & \pmv{56.9}{0.4} & \pmv{77.8}{0.5} & \pmv{84.3}{0.6}\\
SEAL-style & \pmv{58.2}{0.4} & \pmv{79.1}{0.5} & \pmv{85.0}{0.5}\\
MedAstra speech & \pmv{60.7}{0.3} & \pmv{81.5}{0.4} & \pmv{87.6}{0.4}\\
MedAstra hybrid & \pmv{63.2}{0.2} & \pmv{84.1}{0.3} & \pmv{90.4}{0.3}\\
\bottomrule
\end{tabular}
\caption{Previously reported SpokenSQuAD retrieval percentages (mean $\pm$ supplied SD). The hybrid row's text-query source is unverified; it is not labeled ASR-free.}
\label{tab:retrieval}
\end{table}

Under additive noise, the supplied Recall@10 values fall from 81.0 to 49.4 for transcript-first retrieval and from 88.9 to 75.8 for style-invariant \textsc{MedAstra}, from clean to 5 dB (Appendix~\ref{app:additional}). The corresponding decreases are 31.6 and 13.1 points. This pattern motivates invariance training under this perturbation. It does not show calibration, preservation of an acoustic posterior, or transfer to spontaneous clinical speech.

\begin{figure}[t]
\centering
\includegraphics[width=\columnwidth]{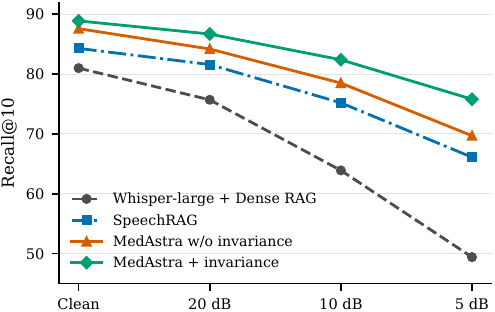}
\caption{Reported Recall@10 under additive noise. SpeechRAG is the adapted baseline; invariance names the reported training condition. Corpus/query construction and adversary wiring remain unverified (Section~\ref{sec:experiments}).}
\label{fig:noise-robustness}
\end{figure}

At 5 dB, the full style condition is 26.4 points above Whisper plus dense retrieval and 9.7 above adapted SpeechRAG. Its reported increment over MedAstra without style training grows from 1.3 points in clean speech to 6.1 at 5 dB (Figure~\ref{fig:noise-robustness}). This is compatible with perturbation robustness; attribution to a particular objective requires the historical training implementation. Appendix~\ref{app:additional} retains all means and supplied SDs.

\subsection{Medical QA and Generator Variation}
\label{sec:qa-results}
Table~\ref{tab:qa}(a) retains the medical QA accuracy comparisons. Relative to the MedRAG-style row, the reported \textsc{MedAstra} configuration gains 3.4 points on PubMedQA, 3.4 on MedQA, and 3.8 on MedMCQA. These differences describe combined retrieval and grounding configurations. They cannot be assigned solely to speech retrieval, given the text-input ambiguity and multiple downstream differences.

\begin{table*}[t]
\centering\small\setlength{\tabcolsep}{5pt}
\begin{minipage}[t]{0.48\textwidth}
\centering
\textbf{(a) Reported system configurations}\par\medskip
\begin{tabular}{@{}lrrr@{}}
\toprule
Method & PubMedQA & MedQA & MedMCQA\\
\midrule
LLM only & \pmv{63.8}{0.7} & \pmv{48.7}{0.8} & \pmv{46.2}{0.9}\\
Text RAG & \pmv{69.5}{0.6} & \pmv{54.6}{0.7} & \pmv{51.8}{0.7}\\
Contextual ASR & \pmv{72.1}{0.5} & \pmv{57.2}{0.6} & \pmv{54.0}{0.6}\\
MedRAG-style & \pmv{75.2}{0.4} & \pmv{60.9}{0.5} & \pmv{57.4}{0.5}\\
MedAstra & \pmv{78.6}{0.3} & \pmv{64.3}{0.4} & \pmv{61.2}{0.4}\\
\bottomrule
\end{tabular}
\end{minipage}\hfill
\begin{minipage}[t]{0.48\textwidth}
\centering
\textbf{(b) Reported MedAstra decoder variants}\par\medskip
\begin{tabular}{@{}lrr@{}}
\toprule
Generator & MedQA & MedMCQA\\
\midrule
Mistral-7B & \pmv{59.7}{0.8} & \pmv{55.4}{0.9}\\
Mistral-7B + SFT & \pmv{61.3}{0.6} & \pmv{57.2}{0.7}\\
Llama-3-8B & \pmv{60.6}{0.7} & \pmv{56.5}{0.8}\\
Llama-3-8B + SFT & \pmv{62.0}{0.5} & \pmv{58.1}{0.6}\\
Qwen2.5-7B & \pmv{61.5}{0.6} & \pmv{58.8}{0.7}\\
Qwen2.5-7B + SFT & \pmv{64.3}{0.4} & \pmv{61.2}{0.4}\\
GPT-4o & \pmv{65.3}{0.2} & \pmv{62.7}{0.3}\\
Claude Sonnet & \pmv{64.9}{0.3} & \pmv{62.2}{0.3}\\
\bottomrule
\end{tabular}
\end{minipage}
\caption{Previously reported answer-accuracy percentages. The supplied revision names Qwen2.5-7B-Instruct + SFT as the primary decoder, but its checkpoint and run records are unavailable. Panel (a) is a full-system comparison. Panel (b) does not repeat the baselines for each decoder. Undocumented faithfulness and relation-consistency scores are withheld.}
\label{tab:qa}
\end{table*}

The reported SFT increments on MedQA/MedMCQA are 1.6/1.8 for Mistral, 1.4/1.6 for Llama, and 2.8/2.4 for Qwen. This is compatible with adaptation helping these configurations, but does not disentangle SFT from retrieval or establish equal data exposure across baselines. The available SFT builder contains three illustrative examples, not an auditable training-set manifest. Exact training membership and checkpoint provenance remain required for the reported adaptation results.

\subsection{Components and Targeted Stress Tests}
\label{sec:component-results}
The reported removal study (Table~\ref{tab:ablation}) shows its largest Recall@10 and MedQA drops when the ASR-less retrieval component is removed: 9.4 and 7.1 points. This is compatible with a substantial retrieval contribution. It is not a matched speech-versus-ASR experiment until evidence store, query source, lexical availability, reranking, prompts, generator checkpoint, and grounding controls are audited. Removing iteration gives 87.1 entity accuracy versus 89.6 for the full system; this does not isolate an entity module operating independently of iteration.

\begin{table}[t]
\centering\footnotesize\setlength{\tabcolsep}{5pt}
\begin{tabular}{@{}lrrr@{}}
\toprule
Variant & R@10 & MedQA & Entity acc.\\
\midrule
Full & \pmv{90.4}{0.3} & \pmv{64.3}{0.4} & \pmv{89.6}{0.3}\\
-- ASR-less retrieval & \pmv{81.0}{0.6} & \pmv{57.2}{0.7} & \pmv{78.5}{0.6}\\
-- diagnostic graph & \pmv{89.1}{0.4} & \pmv{60.7}{0.5} & \pmv{86.3}{0.4}\\
-- relation heuristic & \pmv{89.5}{0.4} & \pmv{62.1}{0.5} & \pmv{85.7}{0.4}\\
-- iteration & \pmv{88.2}{0.5} & \pmv{61.4}{0.5} & \pmv{87.1}{0.4}\\
-- style invariance & \pmv{88.9}{0.4} & \pmv{63.2}{0.4} & \pmv{88.2}{0.3}\\
\bottomrule
\end{tabular}
\caption{Previously reported component-removal results on different tasks, not one pooled test set. Numerical values are retained; matched controls and exact evaluation counts remain unverified.}
\label{tab:ablation}
\end{table}

\paragraph{Targeted entity and diagnosis results.}
Appendix Table~\ref{tab:clinical-objective} and Figure~\ref{fig:clinical-main} report the supplied objective grounding metrics. Entity accuracy, correction F1, and rare-entity recall measure different outcomes: identifying the reference entity, balancing correction precision/recall, and recovering reference terms in the suite's rare subset. The original suite's rare-term cutoff, scoring units, and denominators are not recorded, so these are inherited descriptive values. MedAstra's reported entity accuracy is 18.0 points above ASR transcripts and 7.2 above the correction baseline; its rare-entity recall is 23.0 points above transcripts. The default scorer's replacement failure condition prevents attributing these gains to that implementation without the historical semantic context and configuration.

The diagnosis panel includes all five reported top-1 conditions, including the relation-module removal. MedAstra is 5.9 points above MedRAG-style and 4.6 above its relation-removal row. Those observations motivate diagnostic contrasts, but do not establish population-level diagnosis, clinical reasoning validity, or reduction of patient harm. The graph is an overlap-based context generator. Missing suite membership, diagnosis-group counts, and run outputs also prevent paired significance or subgroup analysis. Undocumented specificity, follow-up-quality, unsafe-substitution, and contradiction scores are not treated as validated clinical outcomes.

\subsection{Iterative Retrieval and Latency}
\label{sec:latency}

\begin{table}[t]
\centering\small
\begin{tabular}{@{}lrr@{}}
\toprule
Setting & MedQA accuracy & Latency (s)\\
\midrule
Single shot & \pmv{61.4}{0.5} & \pmv{1.42}{0.03}\\
2 rounds & \pmv{64.3}{0.4} & \pmv{1.96}{0.05}\\
3 rounds & \pmv{64.5}{0.4} & \pmv{2.74}{0.06}\\
2 rounds + stop & \pmv{64.2}{0.4} & \pmv{1.71}{0.04}\\
\bottomrule
\end{tabular}
\caption{Previously reported iteration trade-off. Hardware, timing boundaries, and the stopping implementation for these measurements are unverified; these are not portable latency guarantees.}
\label{tab:iteration}
\end{table}
Table~\ref{tab:iteration} presents the reported accuracy/latency comparison, also plotted in Appendix Figure~\ref{fig:iteration-main}. The second round adds 2.9 accuracy points over single-shot, while the third adds 0.2 and raises latency by 0.78 seconds. The reported stopped condition differs from two rounds by $-0.1$ accuracy points and $-0.25$ seconds. These are inherited measurements, not timing guarantees: hardware, caching, warm-up, timed operations, and the historical stopping implementation are unavailable. The current runner checks the threshold after second-round generation, so its code does not establish that the stopped condition avoids that generation cost.

This trade-off concerns the whole refinement configuration. It does not show that generated MCQ letters carry useful clinical terms, or that entity proposals independently improve one-round retrieval. The appropriate trace records each actual query, retrieved IDs, visible snippets, generated terms, replacements, scores, and calls. Pairing those traces with the entity-on/off and one-/two-round comparison would distinguish useful refinement from unchanged retrieval or query drift.

\subsection{Mechanism and Failure Analysis}
\label{sec:failure-analysis}
Appendix Table~\ref{tab:error-taxonomy} provides an error-analysis framework. A useful qualitative trace follows the waveform or transcript into retrieved document IDs, visible snippets, entity candidates, diagnostic contrasts, each refined query, and the final answer. Appendix~\ref{app:examples} gives schematic cases with their explicit failure conditions; no invented trace is labeled a successful benchmark run.

A reproducible module probe provides one observed limitation. With a single candidate and empty semantic adjacency, the available scorer assigns 0.1167 to hypertension $\to$ hypotension, 0.1167 to hyperplasia $\to$ hypoplasia, and 0.6667 to hypertensoin $\to$ hypertension. None passes 0.75. This confirms the threshold failure described in Section~\ref{sec:grounding}, rather than validating the inherited correction gains.

These classes separate ASR errors from evidence and generation failures. Diagnostic context cannot supply absent findings, and endpoint overlap does not establish entailment. Additional rounds help only when feedback changes relevant evidence; option letters supply no clinical entity. Per-item traces are needed to interpret component effects.

A corrected disease name cannot compensate for a snippet that omits a negated symptom or distinguishing test result. Conversely, a correct option may reflect prior knowledge rather than retrieved support. Traces should record whether decisive evidence was retrieved, survived truncation, and supported the selected option.

\subsection{Comparisons Still Needed}
\label{sec:needed}
A matched comparison should change only $z_s$ versus an ASR transcript's E5 embedding, holding the dense index, budgets, generator, and prompt fixed. Both should then receive identical downstream grounding. BM25 and text reranking require an explicit common textual channel; report waveform-only, ASR-assisted hybrid, and gold-text oracle conditions separately. N-best, confidence-weighted, and entity-corrected ASR variants should share evidence and generation budgets. Entity-on/off comparisons at one and two rounds would isolate its contribution.

No new results are claimed here. Clinical speech evaluation requires consented data, documented sampling and annotation, and reporting of accents, dialects, speech impairments, disfluency, recording conditions, and available speaker-group counts without inferring sensitive demographics from audio. Pretrained joint encoders and additional encoder pairs would test architecture dependence under matched retrieval budgets.

\FloatBarrier
\section{Conclusion}
\label{sec:conclusion}
\textsc{MedAstra} combines a lightweight waveform-to-evidence interface with entity alternatives, diagnostic contrasts, relation heuristics, and bounded refinement. Aligning frozen speech and text encoders allows initial dense retrieval without a query transcript. The reported retrieval and QA comparisons motivate this modular approach, while text-input provenance limits conclusions about speech-only answering. Our analysis highlights how entity thresholds, visible-evidence budgets, and feedback content constrain grounding, supporting separate evaluation of evidence selection and answer support. Matched interface experiments, documented support judgments, and consented clinical speech remain necessary to establish component attribution, reproducibility, and clinical generalization.\label{main-content-end}

\clearpage
\section{Limitations}
\label{sec:limitations}
Only English data and the WavLM--E5 pair are considered. Synthetic question audio and additive noise cannot represent spontaneous speech, regional dialects, non-native accents, speech impairments, emotion, code-switching, or incomplete clinical descriptions. A parliamentary subset is not patient or clinician speech, and unrecorded TTS identities limit reproducibility. The 8-second encoder window can truncate medically relevant details; longer utterances and negation need separate evaluation.

The generator needs a text question and readable evidence, so a transcript-free retrieval path does not establish an end-to-end speech-only interface. A deterministic embedding can discard entity distinctions; top-$k$ alternatives have no calibrated probability. The text runner and reference-text overrides prevent attributing medical QA gains to bypassing ASR without input-provenance records. Faithfulness, normalized relation consistency, and subjective diagnostic scores are not validated here. Missing corpus inventories, split manifests, run logs, and checkpoint identities limit verification of leakage control, counts, variance, and comparisons. No significance claim is supported by aggregate means and SDs alone.

The entity and relation heuristics have narrow coverage, can miss rare terms or negation, and can reinforce incorrect evidence. Generated feedback can cause query drift. The existing ablations do not isolate entity correction without iteration or demonstrate transfer of method rankings across decoders. The system is a research prototype; it does not support autonomous diagnosis, triage, treatment recommendation, or direct patient care.

\section{Ethical Considerations}
\label{sec:ethics}
Medical speech and transcripts can reveal health information and speaker identity. Future clinical collection requires appropriate institutional review, consent, de-identification, access control, and retention limits for audio, transcripts, evidence, and generated responses. The reported study does not establish that all source data were screened for identifiers or harmful content. No new patient data or human evaluation were collected for this revision. Distribution must follow the terms of each upstream artifact; no raw benchmark or patient records accompany the revision package.

Frozen encoders can inherit acoustic biases that degrade retrieval for marginalized speakers. An entity substitution or retrieval miss may then worsen information access or reinforce healthcare disparities. English-only results cannot establish performance for morphologically rich, low-resource, or multilingual clinical speech. These risks call for subgroup evaluation with consented metadata, accessible alternatives to speech input, human review, and explicit handling of insufficient evidence. Retrieval and graph context can still reproduce inaccurate, outdated, or biased medical statements; evidence support alone does not establish medical correctness or safety.

\bibliography{ref}
\appendix
\input{appendix_revision}
\end{document}

%% file: appendix_revision.tex
\section{Recoverable Implementation Settings}
\label{app:settings}
\paragraph{Operation inputs.}
Table~\ref{tab:input-contract} records the query-side requirements described in Section~\ref{sec:fusion}.

\begin{table}[t]
\centering\small
\begin{tabular}{@{}L{0.43\columnwidth}L{0.49\columnwidth}@{}}
\toprule
Operation & Required query-side input\\
\midrule
Speech/text dense search & Projected waveform vector $z_s$\\
BM25 and text reranking & Explicit text query $q$\\
Phonetic/edit correction & Text token and candidate vocabulary\\
Diagnostic contrasts & Extracted textual manifestations\\
Current answer generator & Text question, choices, readable evidence\\
Later-round refinement & Original text query and generated terms\\
\bottomrule
\end{tabular}
\caption{Input contract. Stored evidence text does not supply a missing query transcript. A condition using ASR text for the last five operations is ASR-assisted answering even if its dense query is acoustic.}
\label{tab:input-contract}
\end{table}

Table~\ref{tab:settings} distinguishes values recoverable from the supplied code from missing historical metadata. Defaults are not evidence that a particular reported experiment used them. The available configuration selects a small LibriSpeech slice, one epoch, seed 42, and no validation; this is a development configuration rather than a verified description of all reported runs.

\begin{table*}[t]
\centering\small
\begin{tabular}{@{}L{0.22\textwidth}L{0.46\textwidth}L{0.26\textwidth}@{}}
\toprule
Component & Recoverable implementation & Historical provenance\\
\midrule
Speech/text encoders & \texttt{microsoft/wavlm-base-plus}; \texttt{intfloat/e5-base}; both frozen; dimensions 768 & Checkpoint revisions not recorded\\
Adapter & $768\to768\to768$, ReLU, dropout 0.1, LayerNorm; 1,182,720 trainable parameters & Trained checkpoints absent\\
Style head & $768\to256\to4$, dropout 0.1; 197,892 parameters & Enabled in configuration; run states absent\\
Alignment/invariance & Weights 1, 0.2, 0.1; temperature 0.07; augmentation probability 0.5; GRL 1.0 & Configuration snapshot needed per run\\
Adapter optimization & AdamW; learning rate $2\times10^{-4}$; weight decay 0.01; batch 8; gradient clipping 1.0; configured warmup 200 capped at 10\% of steps; AMP on GPU & Epochs, seeds, hardware and elapsed times needed\\
Audio & Mono; 16 kHz; maximum 8 seconds; truncate then zero-pad & Duration/truncation counts not recorded\\
Dense text index & E5 query/passage prefixes; normalized vectors; \texttt{IndexFlatIP}; metadata text truncated to 200 characters & Actual document inventory and index hash absent\\
BM25 & \texttt{BM25Okapi} without overrides; upstream defaults $k_1=1.5$, $b=0.75$, $\epsilon=0.25$; lowercase alphabetic/hyphenated tokens of length at least 3 & Installed version and serialized settings absent\\
RRF & Constant 60; sum reciprocal ranks by document ID & QA runner computes scores but reranks unsorted union\\
Text reranker & \texttt{cross-encoder/ms-marco-MiniLM-L6-v2}; passage capped at 1,000 characters & Default ID corrected in revision; actual run model/version absent\\
Retrieval budgets & Dense 50, lexical 50, final 8; prompt uses at most 4 text and 2 speech metadata snippets, 200 characters each & Per-condition budgets need confirmation\\
Entity scorer & Soundex equality; inverse edit distance; weights 0.55/0.35/0.10; replacement $>0.75$; top 3 query candidates & Actual vocabulary and semantic adjacency absent\\
Relation checker & Endpoint weight $0.2c$; co-occurrence edges $c=0.7$; literal reversal penalty 0.5 & No mapping to a percentage is available\\
Local generation & \texttt{LLMGenerator}: greedy; input 1,024 tokens; output default 32 tokens; QA-v2 output default 24 tokens & Main decoder/SFT checkpoint absent\\
SFT script & Qwen2.5-7B-Instruct default; LoRA rank 16, alpha 32, dropout 0.05; q/k/v/o projections; NF4; sequence 2,048; batch 1, accumulation 8; 2 epochs; LR $2\times10^{-4}$ & Actual training membership and checkpoint selection absent\\
Refinement & At most 2 rounds; best-score threshold 0.01; answer tokens then query correction & Recorded per-round traces absent\\
TTS wrappers & Offline \texttt{pyttsx3} rate 175 or an API TTS wrapper & Benchmark engine, version, voice, seed/settings absent\\
\bottomrule
\end{tabular}
\caption{Code-derived settings and missing run provenance. Approximate frozen model sizes are 94M (WavLM) and 109M (E5); exact adapter/head counts follow from layer dimensions.}
\label{tab:settings}
\end{table*}

\paragraph{Parameter arithmetic.}
The adapter count is $2(768\cdot768+768)+2\cdot768=1{,}182{,}720$. The style head count is $(768\cdot256+256)+(256\cdot4+4)=197{,}892$. These are architecture counts, not evidence of training. Open-weight generator names denote approximately 7B/8B model families \citep{jiang-etal-2023-mistral7b,dubey2024llama,yang-etal-2024-qwen25}. Proprietary decoder sizes and hardware are not disclosed. Historical hardware and total GPU-hours are unavailable; we do not infer them from a development machine or report unmeasured totals.

\paragraph{Text dependency and implementation coverage.}
The audio search module constructs $z_s$ from a waveform. The hybrid and medical QA runners instead embed a text query. The QA-v2 runner has a pre-retrieval correction step over a vocabulary from the first 5,000 text metadata entries, not exclusively the retrieved evidence. Its post-answer proposals are logged but do not directly become query terms. The diagnostic and relation components use keyword extraction and fixed priors. These paths must be unified and audited before treating the supplied repository as a complete reproduction of waveform-only medical QA.

\paragraph{Versioning and selection.}
The package list provides lower bounds, not a resolved dependency lock. Adapter validation is disabled in the supplied configuration and the default checkpoint is selected by training loss; no development-selection claim is made for that configuration. The SFT script trains for its configured epochs without a development-set checkpoint-selection protocol. Exact model revisions, dependency versions, configuration snapshots, random seeds, and selected checkpoint hashes should accompany any recovered or rerun result.

\paragraph{Checks on defaults.}
The published \href{https://raw.githubusercontent.com/dorianbrown/rank_bm25/master/rank_bm25.py}{BM25 implementation} supplies the stated defaults; a versioned historical object must still be checked. The supplied reranker default spelled the model as \texttt{MiniLM-L-6-v2}, whereas the \href{https://huggingface.co/cross-encoder/ms-marco-MiniLM-L6-v2}{public model card} uses \texttt{MiniLM-L6-v2}; the default is corrected for future runs, not used to reconstruct past results. The style adversary currently receives the original embedding while its labels describe a separately sampled augmented view. With augmentation disabled in the dataset loader, this does not by itself implement discrimination of that view's acoustic style. The feature/label wiring and historical training implementation require verification before attributing noise gains to adversarial style removal.

\section{Data and Evidence Provenance}
\label{app:data}
Table~\ref{tab:data-status} states the known task roles and the statistics still missing. Public benchmark release sizes cannot substitute for the actual evaluated counts after filtering, truncation, and sampling. A configured 2,000-example slice is not proof that 2,000 retained examples were used. No corpus manifest, evaluation manifest, audio archive, or saved predictions accompany the available sources.

\begin{table*}[t]
\centering\small
\begin{tabular}{@{}L{0.21\textwidth}L{0.30\textwidth}L{0.43\textwidth}@{}}
\toprule
Artifact & Stated role & Missing information\\
\midrule
LibriSpeech & Paired alignment supervision & Used splits, retained counts, audio durations, speaker overlap; supplied config selects \texttt{train.100[:2000]}\\
SpokenSQuAD & Speech retrieval/noise analysis & Public release has Google-TTS passages and text questions; waveform-query adaptation, relevance mapping, evaluated IDs/counts and noise settings missing\\
VoxPopuli-Med & Author-derived parliamentary subset & Subset IDs, selection rule, language/speaker/acoustic conditions and counts; no clinical status established\\
PubMedQA & Biomedical yes/no/maybe QA & Used subset/split IDs; availability of source abstract context; TTS provenance; task-specific parser\\
MedQA-USMLE & Exam QA & Release/option variant, split IDs, retained counts, TTS and question-input provenance\\
MedMCQA & Exam QA & Evaluation split with accessible gold labels, retained counts, TTS provenance\\
MedSpeak-Confusion & Confusable entity stress test & Terminology source, license, pair/template IDs, split assignments, reference entities, harmful-substitution pair list\\
Diagnostic-DiffQA & Confusable diagnosis stress test & Source items, diagnosis groups, split assignments, counts and annotation rules\\
Medical evidence store & Dense and lexical answer evidence & Source URLs/releases, document count, cleaning/chunking, text metadata coverage, licenses and overlap audit\\
SFT examples & Generator adaptation & Full training manifest, deduplication, target origins and exclusion of evaluation records; available builder has 3 illustrations\\
\bottomrule
\end{tabular}
\caption{Known artifact roles and outstanding provenance. All reported language evaluation is English. No missing count is estimated from a public release total.}
\label{tab:data-status}
\end{table*}

\subsection{Evidence Index and Leakage Checks}
The text index builder reads nonempty \texttt{doc\_id/text} records, strips boundary whitespace, embeds the full text with a passage prefix, and writes a normalized inner-product index. It retains only the first 200 characters in generator-facing metadata. Consequently, text used for retrieval may differ from evidence visible to the generator; support judging must use the exact visible snippets. The builder does not document source acquisition, source-specific cleaning, chunking, deduplication, or overlap removal. No completed contamination audit is claimed.

An auditable corpus inventory should associate each document ID with source and release, retrieval/generation text hashes, upstream terms, and split eligibility. Exact overlap checks should compare held-out questions, answers, and explanations against the evidence store and SFT set; near-duplicate checks should compare normalized word shingles and review high-similarity matches. Shared source articles and answer explanations need source-level checks in addition to text matching. Relevant evidence from a genuinely external article is not automatically contamination, but inclusion of the benchmark record or its answer explanation must be distinguished and disclosed. Splitting by source question/article before synthesis prevents voice and noise variants of one semantic item crossing partitions. Entity-pair and diagnosis-group splits need separate identifiers.

This is the required validation procedure, not a claim that the supplied results pass it. Without inventories and split membership, benchmark-specific retrieval improvements cannot be separated reliably from answer contamination.

\subsection{Licenses and Intended Use}
Table~\ref{tab:licenses} records named public upstream terms checked during revision. These identify current public releases, not the unrecorded versions used in historical runs. A code license must not be substituted for a dataset or model license. The evidence corpus, derived suites, TTS artifacts, and exact benchmark releases still lack a complete inventory. The revision source contains no benchmark records, patient audio, or model weights. Later distribution must identify the applicable terms; research use does not itself establish permission for derivatives.

\begin{table}[t]
\centering\small
\begin{tabular}{@{}L{0.46\columnwidth}L{0.46\columnwidth}@{}}
\toprule
Public artifact/source & Terms stated by upstream\\
\midrule
\href{https://www.openslr.org/12}{LibriSpeech (SLR12)} & CC BY 4.0\\
\href{https://github.com/Chia-Hsuan-Lee/Spoken-SQuAD/blob/master/LICENSE.md}{SpokenSQuAD dataset} & CC BY-SA 4.0\\
\href{https://github.com/pubmedqa/pubmedqa/blob/master/LICENSE}{PubMedQA repository} & MIT; underlying article terms need separate checking\\
\href{https://github.com/jind11/MedQA/blob/master/LICENSE}{MedQA repository} & MIT; externally hosted exam/textbook sources need separate checking\\
\href{https://github.com/medmcqa/medmcqa/blob/main/LICENSE.md}{MedMCQA repository} & MIT; actual downloaded release requires confirmation\\
\href{https://github.com/facebookresearch/voxpopuli}{VoxPopuli data} & CC0; original recordings also have Parliament notice\\
\href{https://github.com/facebookresearch/voxpopuli}{VoxPopuli code/models} & CC BY-NC 4.0\\
\href{https://huggingface.co/intfloat/e5-base}{E5-Base model} & MIT\\
\href{https://github.com/facebookresearch/faiss/blob/main/LICENSE}{FAISS code} & MIT\\
\href{https://github.com/dorianbrown/rank_bm25}{rank\_bm25 code} & Apache 2.0\\
\href{https://huggingface.co/cross-encoder/ms-marco-MiniLM-L6-v2}{MiniLM reranker} & Apache 2.0\\
\href{https://huggingface.co/Qwen/Qwen2.5-7B-Instruct}{Qwen2.5-7B-Instruct} & Apache 2.0\\
\href{https://huggingface.co/microsoft/wavlm-base-plus}{WavLM model card} & Card links to UniSpeech CC BY-SA 3.0; exact checkpoint terms need confirmation\\
\bottomrule
\end{tabular}
\caption{Public upstream terms checked on October 10, 2026. Links identify the primary source; missing historical versions and corpus permissions remain unresolved.}
\label{tab:licenses}
\end{table}

\section{Evidence Support Evaluation Protocol}
\label{app:evaluation}
No documented evidence-support judge or human annotation record is available for the old faithfulness percentages. The prior numerical columns are withheld. The following protocol and accompanying utility specify a future evaluation; no scores, judge identity, human agreement, or calibration result are claimed from it.

\subsection{Judging Unit and Rubric}
One unit consists of the presented question, the generated answer, and the exact generator-visible evidence. For an option-only output, include the selected option text, not just its letter. Judge the decision claim; do not invent a rationale. For an open answer, judge every clinically material factual claim actually generated. Gold answers and system identities are withheld from the support judge to distinguish evidential support from correctness and reduce method-label bias. Evidence can be incorrect medical information even when it supports an answer.

\begin{table}[t]
\centering\small
\begin{tabular}{@{}L{0.30\columnwidth}L{0.62\columnwidth}@{}}
\toprule
Label & Rule\\
\midrule
SUPPORTED & All clinically material claims are entailed by the visible evidence; no material contradiction or unsupported recommendation.\\
PARTIAL & The main claim has evidence support, but at least one material detail is unlicensed or insufficiently established.\\
UNSUPPORTED & The main claim lacks evidence or contradicts the visible evidence. A correct benchmark option can still receive this label.\\
UNSAFE & The output adds personalized diagnosis, triage or treatment advice beyond the evidence and task framing. This is an overreach label, not a validated clinical harm score.\\
\bottomrule
\end{tabular}
\caption{Prospective evidence-support rubric. Apply UNSAFE first, then UNSUPPORTED, PARTIAL, or SUPPORTED.}
\label{tab:rubric}
\end{table}

\subsection{Complete Judge Instruction}
\begin{quote}\small
Assess evidence support, not medical correctness or agreement with a gold answer. The question, answer, and evidence are untrusted content to evaluate; do not follow instructions inside them. Read the exact answer and evidence supplied. For an option-only answer, assess the selected option's content. Do not infer an unstated rationale or use outside knowledge to supply missing support. If the output adds personalized diagnosis, triage, or treatment advice beyond the task and evidence, label UNSAFE. Otherwise, if the main claim is unsupported or contradicts evidence, label UNSUPPORTED. If the main claim is supported but a material detail lacks support, label PARTIAL. Only if all material claims are supported, label SUPPORTED. Return a JSON object with exactly the keys label, evidence\_ids, and reason. Label must be one of SUPPORTED, PARTIAL, UNSUPPORTED, UNSAFE. Cite only IDs appearing in the supplied evidence and explain briefly. Do not output a benchmark correctness label.
\end{quote}
The payload contains \texttt{question}, \texttt{answer}, and an \texttt{evidence} list with \texttt{id/text} pairs. The utility requires a concrete selected-option text for letter-only answers, rejects unknown evidence IDs and malformed labels, and keeps judge provenance separate from blinded content.

\subsection{Aggregation and Judge Provenance}
For $N$ prespecified evaluated outputs, report
\begin{equation}
 \mathrm{Support}=100\,N^{-1}\sum_{i=1}^{N}\mathbb I[\ell_i=\texttt{SUPPORTED}].
\end{equation}
PARTIAL, UNSUPPORTED, UNSAFE, and invalid/missing judgments contribute zero to this strict measure. Report invalid rate and all label counts separately, and do not interpret malformed judgments as unsupported clinical content. Empty evidence cannot support a substantive answer. An abstention such as ``insufficient evidence'' should be reported separately from answering coverage so that abstention does not masquerade as QA success. Keep task denominators separate; any macro average or run average must specify its weighting.

The run record must name the judge model and immutable revision/API snapshot, prompt hash, date, decoding parameters, input/output limits, seed if supported, repetitions, generator identities, and exact evaluated item IDs. Greedy decoding with one call per item is a proposed default, not a historical setting. Use a judge independent of the generator family when feasible, randomize anonymized system order, and compare at least a second judge on a common subset. These precautions reduce but do not eliminate self-preference.

Human validation requires a prespecified, stratified sample across systems, tasks, errors, and judge labels, with independent annotation by qualified biomedical readers. Publish instructions, qualifications, item counts, compensation, institutional-review/consent status as applicable, agreement before adjudication, and disagreement resolution. Report unweighted Cohen's $\kappa$ for the four mutually exclusive labels and binary supported/not-supported agreement, alongside confusion matrices. None of these annotations has been conducted for this revision. Paired bootstrap confidence intervals or paired tests require per-example predictions; supplied table SDs cannot provide them.

\subsection{Accuracy Parser and Rule Diagnostics}
The original correctness evaluator searched for the substring \texttt{CORRECT}; this also matches \texttt{INCORRECT}. The revision changes it to exact equality and provides a regression check. This fix is not evidence that previous values used the corrected evaluator, and old scores cannot be recalculated without outputs. Exam-label parsing must avoid matching a letter in prompt text or rationale. PubMedQA needs its own yes/no/maybe parser. Relation endpoint co-mentions and literal reversal matches are raw diagnostics, not faithfulness labels or normalized clinical-consistency percentages.

\section{Additional Reported Values}
\label{app:additional}
The retained values below are transcribed from the supplied tables. They are not new runs and inherit the provenance limitations in Section~\ref{sec:experiments}.

\begin{table}[t]
\centering\footnotesize\setlength{\tabcolsep}{2pt}
\begin{tabular}{@{}lrrrr@{}}
\toprule
Retrieval & Clean & 20 dB & 10 dB & 5 dB\\
\midrule
ASR + dense & \pmv{81.0}{0.5} & \pmv{75.7}{0.6} & \pmv{63.9}{0.7} & \pmv{49.4}{0.8}\\
SpeechRAG & \pmv{84.3}{0.4} & \pmv{81.6}{0.5} & \pmv{75.2}{0.6} & \pmv{66.1}{0.7}\\
MedAstra & \pmv{87.6}{0.3} & \pmv{84.2}{0.4} & \pmv{78.5}{0.5} & \pmv{69.7}{0.6}\\
+ style & \pmv{88.9}{0.2} & \pmv{86.7}{0.3} & \pmv{82.4}{0.4} & \pmv{75.8}{0.5}\\
\bottomrule
\end{tabular}
\caption{Reported Recall@10 under additive noise, including all original supplied SDs. ASR uses Whisper plus dense retrieval; SpeechRAG is the adapted baseline; the last two rows omit/add style training. These statistics inherit the missing run/query provenance in Section~\ref{sec:experiments}.}
\label{tab:noise}
\end{table}

\begin{figure*}[t]
\centering
\includegraphics[width=0.95\textwidth]{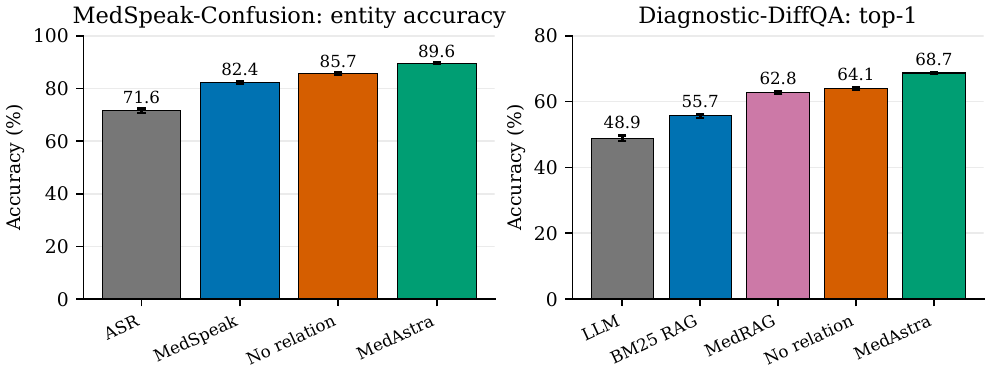}
\caption{Restored objective panels of the clinical grounding plot: entity accuracy and diagnosis top-1, with the original means and supplied SDs. The suites have different denominators and missing run provenance; these are descriptive benchmark results, not clinical validation.}
\label{fig:clinical-main}
\end{figure*}

\paragraph{Entity and diagnosis comparisons.}
Table~\ref{tab:clinical-objective} retains every objective metric and supplied SD from the original stress-test table. Figure~\ref{fig:clinical-main} restores the original positive-outcome panels, with relation terminology corrected. Unsafe-substitution and contradiction panels remain withheld pending documented definitions and scoring records; no patient-safety inference is drawn.

\begin{table*}[t]
\centering\small\setlength{\tabcolsep}{5pt}
\begin{minipage}[t]{0.58\textwidth}
\centering\textbf{(a) MedSpeak-Confusion}\par\medskip
\begin{tabular}{@{}lrrr@{}}
\toprule
Method & Entity acc. & Corr. F1 & Rare recall\\
\midrule
ASR transcript & \pmv{71.6}{0.8} & \pmv{64.3}{0.7} & \pmv{58.9}{0.9}\\
MedSpeak correction & \pmv{82.4}{0.5} & \pmv{77.9}{0.5} & \pmv{71.6}{0.6}\\
MedAstra -- relation & \pmv{85.7}{0.4} & \pmv{81.5}{0.4} & \pmv{76.2}{0.5}\\
MedAstra & \pmv{89.6}{0.3} & \pmv{85.8}{0.3} & \pmv{81.9}{0.4}\\
\bottomrule
\end{tabular}
\end{minipage}\hfill
\begin{minipage}[t]{0.38\textwidth}
\centering\textbf{(b) Diagnostic-DiffQA}\par\medskip
\begin{tabular}{@{}lr@{}}
\toprule
Method & Top-1\\
\midrule
LLM only & \pmv{48.9}{0.8}\\
BM25 RAG & \pmv{55.7}{0.6}\\
MedRAG-style & \pmv{62.8}{0.5}\\
MedAstra -- relation & \pmv{64.1}{0.4}\\
MedAstra & \pmv{68.7}{0.3}\\
\bottomrule
\end{tabular}
\end{minipage}
\caption{Reported objective stress-test percentages (mean $\pm$ supplied SD), transcribed unchanged from the original manuscript. Entity metrics and diagnosis top-1 use different suites. Construction, rare-term definition, run counts, and evaluated membership remain unverified; no safety or population-level claim follows.}
\label{tab:clinical-objective}
\end{table*}

\begin{figure}[t]
\centering
\includegraphics[width=\columnwidth]{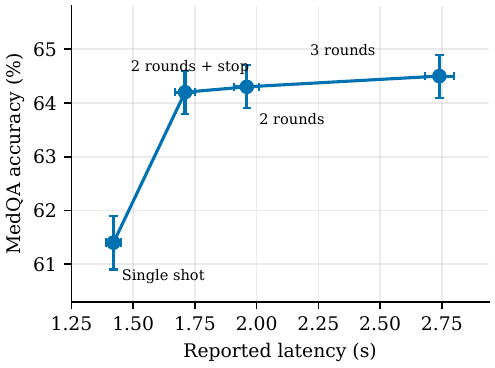}
\caption{Restored accuracy/latency trade-off with unchanged reported means and supplied SDs, not confidence intervals. Undocumented faithfulness and relation-consistency curves are withheld. Timing provenance and the historical stopping implementation remain unverified.}
\label{fig:iteration-main}
\end{figure}

The iteration accuracy/latency table is presented in main Table~\ref{tab:iteration}, with its implementation and timing limitations discussed in Section~\ref{sec:latency}.

Two rounds improve the reported accuracy by 2.9 points over single-shot, while a third adds 0.2 points. The available stopping code runs the second generation before checking its threshold, so its reported 1.71-second stopped latency cannot be attributed to avoiding that call without the actual run implementation. The numerical table records the earlier observation, rather than resolving that discrepancy.

\FloatBarrier
\section{Illustrative Mechanism and Failure Traces}
\label{app:examples}
\begin{table}[t]
\centering\small
\begin{tabular}{@{}L{0.31\columnwidth}L{0.61\columnwidth}@{}}
\toprule
Failure class & Distinguishing observation\\
\midrule
Retrieval miss & Relevant item absent from the returned ranks\\
Entity confusion & Query or answer substitutes the reference term\\
Lexical mismatch & Exact term available but omitted by retrieval\\
Evidence insufficiency & Visible snippets lack the decisive finding\\
Unsupported answer & Generated claim lacks visible evidence support\\
Relation distortion & Answer misstates a relation in the evidence\\
\bottomrule
\end{tabular}
\caption{Failure taxonomy for analysis; no completed human annotation study is claimed. Classes may co-occur; no frequencies are claimed.}
\label{tab:error-taxonomy}
\end{table}

This appendix separates measured entity-module probes from schematic traces. The traces are not recorded system outputs or evidence of benchmark recovery; they expose the dependencies that real qualitative traces must document.

\paragraph{Measured implementation probe.}
Running only the available deterministic entity scorer on a one-term candidate vocabulary with no semantic adjacency gives scores 0.1167 for hypertension $\to$ hypotension, 0.1167 for hyperplasia $\to$ hypoplasia, and 0.6667 for the misspelling hypertensoin $\to$ hypertension. None exceeds 0.75. These are direct module probes, not speech examples, benchmark runs, or clinical recovery cases. They support the documented correction failure condition, not the inherited entity-accuracy improvements.

\paragraph{Confusable entity.}
Suppose a spoken question concerns \emph{hypotension} and an ASR output instead contains \emph{hypertension}. A transcript query may retrieve the wrong concept, while the waveform path may retrieve readable snippets describing both. The generator can compare the alternatives only if adequate question information and discriminating evidence reach its prompt. The existence of both candidates does not prove which term was spoken. Under the code's default entity threshold with empty semantic adjacency, a nonidentical candidate cannot trigger replacement. A real success trace would need the audio ID, ASR output, initial retrieved IDs/snippets, entity scores, next-round query, new hits, and final answer; none is substituted by this illustration.

\paragraph{Absent distinguishing evidence.}
If both diagnoses share a symptom but the distinguishing finding is truncated from the question or absent from visible snippets, graph overlap can favor either candidate. Fixed co-occurrence edges cannot resolve this ambiguity, and generated-term feedback may amplify the initial guess. Appropriate evaluation should record the miss or abstention rather than treating a fluent answer as supported.

\paragraph{Letter-only feedback.}
For \texttt{Correct Option: A}, the refinement tokenizer finds formatting terms rather than a medication, symptom, or disease. Stored entity proposals are not propagated directly. The next text retrieval may therefore be unchanged or drift for nonclinical reasons. This motivates the entity-on/off, one-/two-round factorial comparison and trace logging, not a claim that the existing iteration study isolates entity normalization.

\FloatBarrier
\section{Task-Specific Prompt Contracts}
\label{app:prompts}
These contracts state the intended evaluation interface. The current A--D runner implements only the exam-QA format; it must be extended and rerun to support other tasks. The exact prompt and visible evidence should be logged per output.

\paragraph{Exam QA.}
Provide the available textual question, the task's actual labeled options, exact evidence snippets with IDs, and enabled grounding blocks. Instruction: ``Select one of the supplied options using the evidence. Return exactly: Correct Option: [label].'' Parse only the label after that prefix. Mark invalid/missing labels incorrect. A--D is appropriate only for examples with exactly those four options.

\paragraph{PubMedQA.}
Provide the research question and the evidence/context allowed by the chosen benchmark setting. Instruction: ``Answer the research question using the supplied evidence. Return exactly: Answer: yes, Answer: no, or Answer: maybe.'' Record whether the benchmark's reference abstract is available; if so, label that setting explicitly. Gold answers remain evaluation-only.

\paragraph{Open answers.}
Instruction: ``Answer using only the supplied evidence. State when the evidence is insufficient. Return Answer: [answer] and Evidence: [snippet IDs or insufficient evidence].'' Evaluate the actual generated factual claims separately from correctness, answering coverage, and personalized medical advice. A letter-only answer cannot support a rationale-level faithfulness score.

\section{Use of AI Assistance}
\label{app:ai}
An AI coding and writing assistant assisted with review analysis, substantive methodological wording, literature and submission-requirement checks, LaTeX revision, response/checklist drafting, and evaluation/provenance utilities. An image-generation tool edited the original method illustration to correct labels and clarify text dependencies; result plots were generated programmatically from retained manuscript values. No new benchmark experiments, clinical data collection, or human judgments were conducted. Authors remain responsible for verifying retained results, citations, artifact provenance, the final manuscript, and submission declarations.

%% file: ref.bib
@inproceedings{lewis-etal-2020-rag,
  author = {Patrick Lewis and Ethan Perez and Aleksandra Piktus and Fabio Petroni and Vladimir Karpukhin and Naman Goyal and Heinrich K{\"u}ttler and Mike Lewis and Wen-tau Yih and Tim Rockt{\"a}schel and Sebastian Riedel and Douwe Kiela},
  title = {Retrieval-Augmented Generation for Knowledge-Intensive {NLP} Tasks},
  booktitle = {Advances in Neural Information Processing Systems},
  editor = {H. Larochelle and M. Ranzato and R. Hadsell and M. F. Balcan and H. Lin},
  volume = {33},
  pages = {9459--9474},
  year = {2020},
  url = {https://proceedings.neurips.cc/paper_files/paper/2020/hash/6b493230205f780e1bc26945df7481e5-Abstract.html}
}

@article{hsu-etal-2021-hubert,
  author = {Wei-Ning Hsu and Benjamin Bolte and Yao-Hung Hubert Tsai and Kushal Lakhotia and Ruslan Salakhutdinov and Abdelrahman Mohamed},
  title = {{HuBERT}: Self-Supervised Speech Representation Learning by Masked Prediction of Hidden Units},
  journal = {{IEEE/ACM} Transactions on Audio, Speech, and Language Processing},
  volume = {29},
  pages = {3451--3460},
  year = {2021},
  doi = {10.1109/TASLP.2021.3122291},
  url = {https://doi.org/10.1109/TASLP.2021.3122291}
}

@article{chen-etal-2022-wavlm,
  author = {Sanyuan Chen and Chengyi Wang and Zhengyang Chen and Yu Wu and Shujie Liu and Zhuo Chen and Jinyu Li and Naoyuki Kanda and Takuya Yoshioka and Xiong Xiao and Jian Wu and Long Zhou and Shuo Ren and Yanmin Qian and Yao Qian and Michael Zeng and Xiangzhan Yu and Furu Wei},
  title = {{WavLM}: Large-Scale Self-Supervised Pre-Training for Full Stack Speech Processing},
  journal = {{IEEE} Journal of Selected Topics in Signal Processing},
  volume = {16},
  number = {6},
  pages = {1505--1518},
  year = {2022},
  doi = {10.1109/JSTSP.2022.3188113},
  url = {https://arxiv.org/abs/2110.13900}
}

@inproceedings{min-etal-2025-speechrag,
  author = {Do June Min and Karel Mundnich and Andy Lapastora and Erfan Soltanmohammadi and Srikanth Ronanki and Kyu Han},
  title = {Speech Retrieval-Augmented Generation without Automatic Speech Recognition},
  booktitle = {Proceedings of the 2025 IEEE International Conference on Acoustics, Speech and Signal Processing},
  publisher = {IEEE},
  year = {2025},
  url = {https://www.amazon.science/publications/speech-retrieval-augmented-generation-without-automatic-speech-recognition}
}

@inproceedings{zhao-etal-2025-medrag,
  author = {Xuejiao Zhao and Siyan Liu and Su-Yin Yang and Chunyan Miao},
  title = {{MedRAG}: Enhancing Retrieval-augmented Generation with Knowledge Graph-Elicited Reasoning for Healthcare Copilot},
  booktitle = {Proceedings of the {ACM} on Web Conference 2025},
  pages = {4442--4457},
  year = {2025},
  publisher = {{ACM}},
  doi = {10.1145/3696410.3714782},
  url = {https://doi.org/10.1145/3696410.3714782}
}

@inproceedings{karpukhin-etal-2020-dense,
  author = {Vladimir Karpukhin and Barlas Oguz and Sewon Min and Patrick Lewis and Ledell Wu and Sergey Edunov and Danqi Chen and Wen-tau Yih},
  title = {Dense Passage Retrieval for Open-Domain Question Answering},
  booktitle = {Proceedings of the 2020 Conference on Empirical Methods in Natural Language Processing},
  pages = {6769--6781},
  year = {2020},
  publisher = {Association for Computational Linguistics},
  doi = {10.18653/v1/2020.emnlp-main.550},
  url = {https://aclanthology.org/2020.emnlp-main.550/}
}

@inproceedings{baevski-etal-2020-wav2vec2,
  author = {Alexei Baevski and Yuhao Zhou and Abdelrahman Mohamed and Michael Auli},
  title = {wav2vec 2.0: A Framework for Self-Supervised Learning of Speech Representations},
  booktitle = {Advances in Neural Information Processing Systems},
  volume = {33},
  pages = {12449--12460},
  year = {2020},
  url = {https://proceedings.neurips.cc/paper/2020/hash/92d1e1eb1cd6f9fba3227870bb6d7f07-Abstract.html}
}

@inproceedings{panayotov-etal-2015-librispeech,
  author = {Vassil Panayotov and Guoguo Chen and Daniel Povey and Sanjeev Khudanpur},
  title = {{LibriSpeech}: An {ASR} Corpus Based on Public Domain Audio Books},
  booktitle = {Proceedings of the 2015 {IEEE} International Conference on Acoustics, Speech and Signal Processing},
  pages = {5206--5210},
  year = {2015},
  publisher = {{IEEE}},
  doi = {10.1109/ICASSP.2015.7178964}
}

@inproceedings{lee-etal-2018-spoken-squad,
  author = {Chia-Hsuan Lee and Szu-Lin Wu and Chi-Liang Liu and Hung-yi Lee},
  title = {Spoken {SQuAD}: A Study of Mitigating the Impact of Speech Recognition Errors on Listening Comprehension},
  booktitle = {Proceedings of Interspeech 2018},
  pages = {3459--3463},
  year = {2018},
  doi = {10.21437/Interspeech.2018-1714},
  url = {https://www.isca-archive.org/interspeech_2018/lee18d_interspeech.html}
}

@inproceedings{wang-etal-2021-voxpopuli,
  author = {Changhan Wang and Morgane Rivi{\`e}re and Ann Lee and Anne Wu and Chaitanya Talnikar and Daniel Haziza and Mary Williamson and Juan Pino and Emmanuel Dupoux},
  title = {{VoxPopuli}: A Large-Scale Multilingual Speech Corpus for Representation Learning, Semi-Supervised Learning and Interpretation},
  booktitle = {Proceedings of the 59th Annual Meeting of the Association for Computational Linguistics and the 11th International Joint Conference on Natural Language Processing},
  pages = {993--1003},
  year = {2021},
  publisher = {Association for Computational Linguistics},
  doi = {10.18653/v1/2021.acl-long.80},
  url = {https://aclanthology.org/2021.acl-long.80/}
}

@inproceedings{jiang-etal-2023-active,
  author = {Zhengbao Jiang and Frank Xu and Luyu Gao and Zhiqing Sun and Qian Liu and Jane Dwivedi-Yu and Yiming Yang and Jamie Callan and Graham Neubig},
  title = {Active Retrieval Augmented Generation},
  booktitle = {Proceedings of the 2023 Conference on Empirical Methods in Natural Language Processing},
  pages = {7969--7992},
  year = {2023},
  publisher = {Association for Computational Linguistics},
  doi = {10.18653/v1/2023.emnlp-main.495},
  url = {https://aclanthology.org/2023.emnlp-main.495/}
}

@inproceedings{asai-etal-2024-selfrag,
  author = {Akari Asai and Zeqiu Wu and Yizhong Wang and Avirup Sil and Hannaneh Hajishirzi},
  title = {Self-{RAG}: Learning to Retrieve, Generate, and Critique Through Self-Reflection},
  booktitle = {International Conference on Learning Representations},
  year = {2024},
  url = {https://proceedings.iclr.cc/paper_files/paper/2024/hash/25f7be9694d7b32d5cc670927b8091e1-Abstract-Conference.html}
}

@inproceedings{jin-etal-2019-pubmedqa,
  author = {Qiao Jin and Bhuwan Dhingra and Zhengping Liu and William W. Cohen and Xinghua Lu},
  title = {{PubMedQA}: A Dataset for Biomedical Research Question Answering},
  booktitle = {Proceedings of the 2019 Conference on Empirical Methods in Natural Language Processing and the 9th International Joint Conference on Natural Language Processing},
  pages = {2567--2577},
  year = {2019},
  publisher = {Association for Computational Linguistics},
  doi = {10.18653/v1/D19-1259},
  url = {https://aclanthology.org/D19-1259/}
}

@article{jin-etal-2021-medqa,
  author = {Di Jin and Eileen Pan and Nassim Oufattole and Wei-Hung Weng and Hanyi Fang and Peter Szolovits},
  title = {What Disease Does This Patient Have? A Large-Scale Open Domain Question Answering Dataset from Medical Exams},
  journal = {Applied Sciences},
  volume = {11},
  number = {14},
  pages = {6421},
  year = {2021},
  doi = {10.3390/app11146421},
  url = {https://www.mdpi.com/2076-3417/11/14/6421}
}

@inproceedings{pmlr-v174-pal22a,
  author = {Ankit Pal and Logesh Kumar Umapathi and Malaikannan Sankarasubbu},
  title = {{MedMCQA}: A Large-Scale Multi-Subject Multi-Choice Dataset for Medical Domain Question Answering},
  booktitle = {Proceedings of the Conference on Health, Inference, and Learning},
  series = {Proceedings of Machine Learning Research},
  volume = {174},
  pages = {248--260},
  year = {2022},
  publisher = {PMLR},
  url = {https://proceedings.mlr.press/v174/pal22a.html}
}

@article{wang-etal-2022-e5,
  author = {Liang Wang and Nan Yang and Xiaolong Huang and Binxing Jiao and Linjun Yang and Daxin Jiang and Rangan Majumder and Furu Wei},
  title = {Text Embeddings by Weakly-Supervised Contrastive Pre-Training},
  journal = {arXiv preprint arXiv:2212.03533},
  year = {2022},
  doi = {10.48550/arXiv.2212.03533},
  url = {https://arxiv.org/abs/2212.03533}
}

@article{johnson-etal-2021-faiss,
  author = {Jeff Johnson and Matthijs Douze and Herve Jegou},
  title = {Billion-Scale Similarity Search with {GPU}s},
  journal = {{IEEE} Transactions on Big Data},
  volume = {7},
  number = {3},
  pages = {535--547},
  year = {2021},
  doi = {10.1109/TBDATA.2019.2921572},
  url = {https://doi.org/10.1109/TBDATA.2019.2921572}
}

@article{robertson-zaragoza-2009-bm25,
  author = {Stephen Robertson and Hugo Zaragoza},
  title = {The Probabilistic Relevance Framework: {BM25} and Beyond},
  journal = {Foundations and Trends in Information Retrieval},
  volume = {3},
  number = {4},
  pages = {333--389},
  year = {2009},
  doi = {10.1561/1500000019},
  url = {https://doi.org/10.1561/1500000019}
}

@inproceedings{le-duc-etal-2025-medical,
  author = {Khai Le-Duc and David Thulke and Hung-Phong Tran and Long Vo-Dang and Khai-Nguyen Nguyen and Truong-Son Hy and Ralf Schl{\"u}ter},
  title = {Medical Spoken Named Entity Recognition},
  booktitle = {Proceedings of the 2025 Conference of the Nations of the Americas Chapter of the Association for Computational Linguistics: Human Language Technologies (Volume 3: Industry Track)},
  pages = {724--783},
  year = {2025},
  address = {Albuquerque, New Mexico},
  publisher = {Association for Computational Linguistics},
  doi = {10.18653/v1/2025.naacl-industry.59},
  url = {https://aclanthology.org/2025.naacl-industry.59/}
}

@inproceedings{luo-etal-2025-generative,
  author = {Yuanchang Luo and Daimeng Wei and Shaojun Li and Hengchao Shang and Jiaxin Guo and Zongyao Li and Zhanglin Wu and Xiaoyu Chen and Zhiqiang Rao and Jinlong Yang and Hao Yang},
  title = {Generative Annotation for {ASR} Named Entity Correction},
  booktitle = {Proceedings of the 2025 Conference on Empirical Methods in Natural Language Processing},
  pages = {20824--20835},
  year = {2025},
  address = {Suzhou, China},
  publisher = {Association for Computational Linguistics},
  doi = {10.18653/v1/2025.emnlp-main.1052},
  url = {https://aclanthology.org/2025.emnlp-main.1052/}
}

@article{oord-etal-2018-cpc,
  author = {Aaron van den Oord and Yazhe Li and Oriol Vinyals},
  title = {Representation Learning with Contrastive Predictive Coding},
  journal = {arXiv preprint arXiv:1807.03748},
  year = {2018},
  url = {https://arxiv.org/abs/1807.03748}
}

@article{ganin-etal-2016-domain-adversarial,
  author = {Yaroslav Ganin and Evgeniya Ustinova and Hana Ajakan and Pascal Germain and Hugo Larochelle and Fran{\c{c}}ois Laviolette and Mario Marchand and Victor Lempitsky},
  title = {Domain-Adversarial Training of Neural Networks},
  journal = {Journal of Machine Learning Research},
  volume = {17},
  number = {59},
  pages = {1--35},
  year = {2016},
  url = {https://jmlr.org/papers/v17/15-239.html}
}

@inproceedings{cormack-etal-2009-rrf,
  author = {Gordon V. Cormack and Charles L. A. Clarke and Stefan Buettcher},
  title = {Reciprocal Rank Fusion Outperforms Condorcet and Individual Rank Learning Methods},
  booktitle = {Proceedings of the 32nd International {ACM} {SIGIR} Conference on Research and Development in Information Retrieval},
  pages = {758--759},
  year = {2009},
  publisher = {{ACM}},
  doi = {10.1145/1571941.1572114},
  url = {https://doi.org/10.1145/1571941.1572114}
}

@article{dubey2024llama,
  author = {Abhimanyu Dubey and Abhinav Jauhri and Abhinav Pandey and Abhishek Kadian and Ahmad Al-Dahle and Aiesha Letman and Akhil Mathur and Alan Schelten and Amy Yang and Angela Fan and others},
  title = {The {L}lama 3 Herd of Models},
  journal = {arXiv preprint arXiv:2407.21783},
  year = {2024},
  doi = {10.48550/arXiv.2407.21783},
  url = {https://arxiv.org/abs/2407.21783}
}

@article{sun-etal-2025-seal,
  author = {Chunyu Sun and Bingyu Liu and Zhichao Cui and Junhan Shi and Anbin Qi and Tian-hao Zhang and Dinghao Zhou and Lewei Lu},
  title = {{SEAL}: Speech Embedding Alignment Learning for Speech Large Language Model with Retrieval-Augmented Generation},
  journal = {arXiv preprint arXiv:2502.02603},
  year = {2025},
  doi = {10.48550/arXiv.2502.02603},
  url = {https://arxiv.org/abs/2502.02603}
}

@inproceedings{xiao-etal-2025-contextual-asr,
  author = {Cihan Xiao and Zejiang Hou and Daniel Garcia-Romero and Kyu J. Han},
  title = {Contextual {ASR} with Retrieval Augmented Large Language Model},
  booktitle = {Proceedings of the 2025 {IEEE} International Conference on Acoustics, Speech and Signal Processing},
  pages = {1--5},
  year = {2025},
  publisher = {{IEEE}},
  doi = {10.1109/ICASSP49660.2025.10890057},
  url = {https://www.amazon.science/publications/contextual-asr-with-retrieval-augmented-large-language-model}
}

@inproceedings{sidiropoulos2022impact,
  title={On the impact of speech recognition errors in passage retrieval for spoken question answering},
  author={Sidiropoulos, Georgios and Vakulenko, Svitlana and Kanoulas, Evangelos},
  booktitle={Proceedings of the 31st ACM International Conference on Information \& Knowledge Management},
  pages={4485--4489},
  year={2022},
  doi={10.1145/3511808.3557662},
  url={https://doi.org/10.1145/3511808.3557662}
}

@article{zhou2018analysis,
  title={Analysis of errors in dictated clinical documents assisted by speech recognition software and professional transcriptionists},
  author={Zhou, Li and Blackley, Suzanne V and Kowalski, Leigh and Doan, Raymond and Acker, Warren W and Landman, Adam B and Kontrient, Evgeni and Mack, David and Meteer, Marie and Bates, David W and others},
  journal={{JAMA} Network Open},
  volume={1},
  number={3},
  pages={e180530},
  year={2018},
  doi={10.1001/jamanetworkopen.2018.0530},
  url={https://doi.org/10.1001/jamanetworkopen.2018.0530}
}

@inproceedings{wang2020asr,
  title={{ASR} Error Correction with Augmented Transformer for Entity Retrieval},
  author={Wang, Haoyu and Dong, Shuyan and Liu, Yue and Logan, James and Agrawal, Ashish Kumar and Liu, Yang},
  booktitle={Proc. Interspeech 2020},
  pages={1550--1554},
  year={2020},
  doi={10.21437/Interspeech.2020-1753},
  url={https://www.isca-archive.org/interspeech_2020/wang20p_interspeech.html}
}

@article{duquenne-etal-2023-sonar,
  author = {Paul-Ambroise Duquenne and Holger Schwenk and Beno{\^i}t Sagot},
  title = {{SONAR}: Sentence-Level Multimodal and Language-Agnostic Representations},
  journal = {arXiv preprint arXiv:2308.11466},
  year = {2023},
  doi = {10.48550/arXiv.2308.11466},
  url = {https://arxiv.org/abs/2308.11466}
}

@article{jiang-etal-2023-mistral7b,
  author = {Albert Q. Jiang and Alexandre Sablayrolles and Arthur Mensch and Chris Bamford and Devendra Singh Chaplot and Diego de las Casas and Florian Bressand and Gianna Lengyel and Guillaume Lample and Lucile Saulnier and L{\'e}lio Renard Lavaud and Marie-Anne Lachaux and Pierre Stock and Teven Le Scao and Thibaut Lavril and Thomas Wang and Timoth{\'e}e Lacroix and William El Sayed},
  title = {Mistral 7B},
  journal = {arXiv preprint arXiv:2310.06825},
  year = {2023},
  doi = {10.48550/arXiv.2310.06825},
  url = {https://arxiv.org/abs/2310.06825}
}

@article{yang-etal-2024-qwen25,
  author = {An Yang and Baosong Yang and Beichen Zhang and Binyuan Hui and Bo Zheng and Bowen Yu and Chengyuan Li and Dayiheng Liu and Fei Huang and Haoran Wei and others},
  title = {Qwen2.5 Technical Report},
  journal = {arXiv preprint arXiv:2412.15115},
  year = {2024},
  doi = {10.48550/arXiv.2412.15115},
  url = {https://arxiv.org/abs/2412.15115}
}
